\documentclass[sigconf]{acmart}

\usepackage{tabularx}
\usepackage{balance} 
\usepackage{framed}
\usepackage{xcolor}
\usepackage{multirow}
\definecolor{framecolor}{rgb}{0.8,0.2,0.2} 
 {\endMakeFramed}

\usepackage[T1]{fontenc}
\usepackage[utf8]{inputenc}
\usepackage{CJKutf8}

\usepackage{graphicx} 
\usepackage{listings}
\usepackage{xcolor}

\newcommand{\A}{\raisebox{-0.15\height}{\includegraphics[height=1.2em]{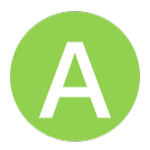}}}

\newcommand{\B}{\raisebox{-0.15\height}{\includegraphics[height=1.2em]{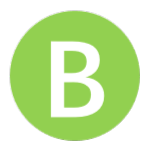}}}

\newcommand{\C}{\raisebox{-0.15\height}{\includegraphics[height=1.2em]{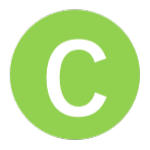}}}

\newcommand{\asmall}{\raisebox{-0.15\height}{\includegraphics[height=1.2em]{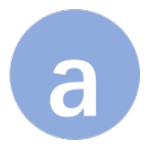}}}

\newcommand{\esmall}{\raisebox{-0.15\height}{\includegraphics[height=1.2em]{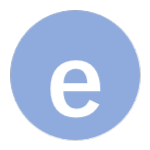}}}

\AtBeginDocument{%
  }

\copyrightyear{2025}
\acmYear{2025}
\setcopyright{cc}
\setcctype{by}
\acmConference[UbiComp Companion '25]{Companion of the the 2025 ACM International Joint Conference on Pervasive and Ubiquitous Computing}{October 12--16, 2025}{Espoo, Finland}
\acmBooktitle{Companion of the the 2025 ACM International Joint Conference on Pervasive and Ubiquitous Computing (UbiComp Companion '25), October 12--16, 2025, Espoo, Finland}\acmDOI{10.1145/3714394.3756343}
\acmISBN{979-8-4007-1477-1/2025/10}

\begin{document}

\title[CTG-Insight: A Multi-Agent Interpretable LLM Framework for Cardiotocography Analysis and Classification]{CTG-Insight: A Multi-Agent Interpretable LLM Framework for Cardiotocography Analysis and Classification}

\author{Black Sun}
\affiliation{%
 \institution{Department of Computer Science \\ Aarhus University}
 \city{Aarhus}
 \country{Denmark}
 }
\email{blackthompson770@gmail.com}

\author{Die (Delia) Hu}
\affiliation{%
 \institution{Anhui University of Science and Technology}
 \city{Anhui}
 \country{China}
 }
\email{die.hu2015@outlook.com}


\renewcommand{\shortauthors}{Black Sun and Die (Delia) Hu}


\begin{abstract}
Remote fetal monitoring technologies are becoming increasingly common. Yet, most current systems offer limited interpretability, leaving expectant parents with raw cardiotocography (CTG) data that is difficult to understand. In this work, we present CTG-Insight, a multi-agent LLM system that provides structured interpretations of fetal heart rate (FHR) and uterine contraction (UC) signals. Drawing from established medical guidelines, CTG-Insight decomposes each CTG trace into five medically defined features: baseline, variability, accelerations, decelerations, and sinusoidal pattern, each analyzed by a dedicated agent. A final aggregation agent synthesizes the outputs to deliver a holistic classification of fetal health, accompanied by a natural language explanation. We evaluate CTG-Insight on the NeuroFetalNet Dataset and compare it against deep learning models and the single-agent LLM baseline. Results show that CTG-Insight achieves state-of-the-art accuracy (96.4\%) and F1-score (97.8\%) while producing transparent and interpretable outputs. This work contributes an interpretable and extensible CTG analysis framework.
\end{abstract}




\maketitle

\section{Introduction}

\begin{figure*}[h!]
    \centering
    \includegraphics[width=1\textwidth]{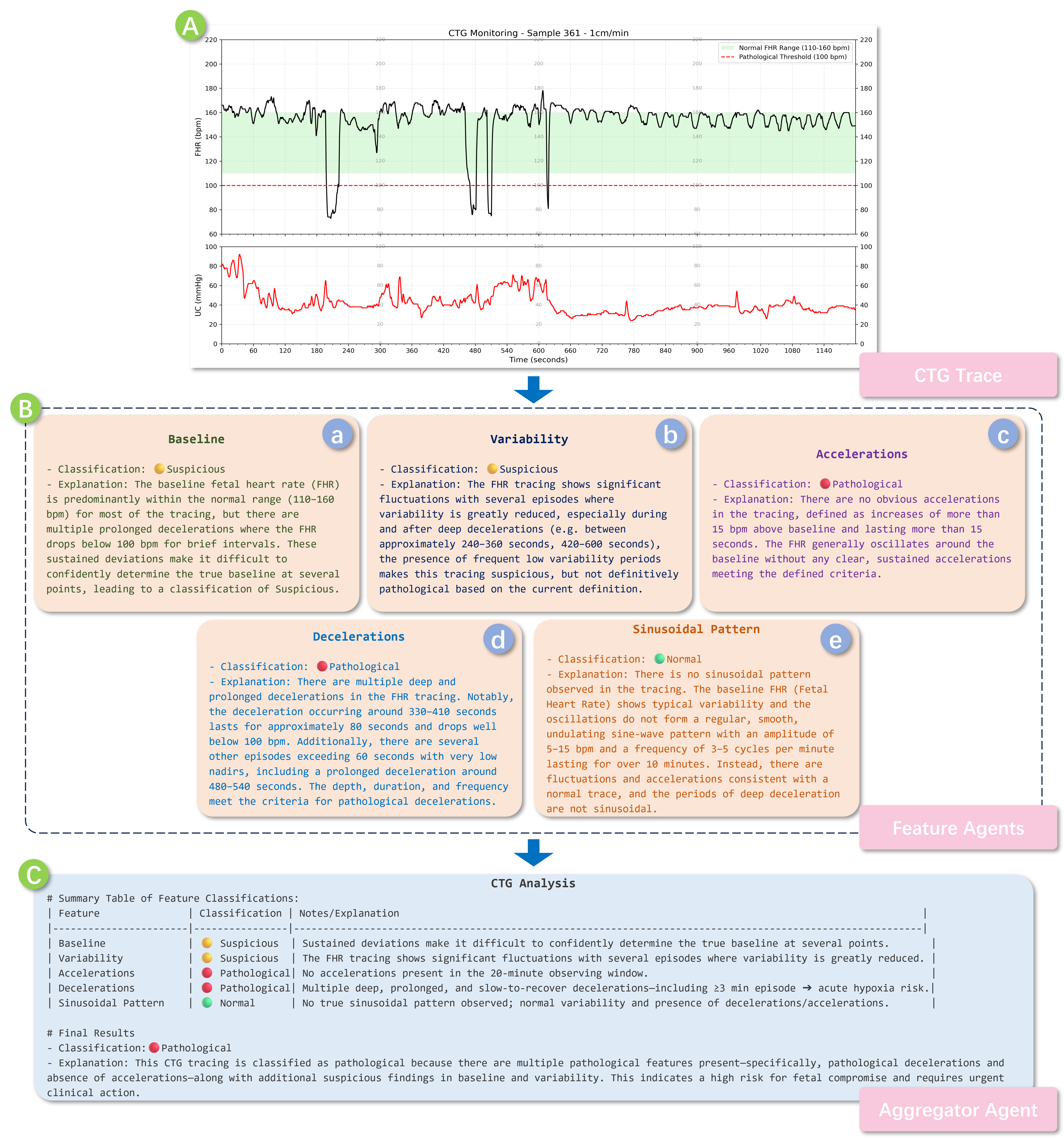}
    \caption{Overall workflow of the CTG-Insight system. \protect\A{} The cardiotocograph (CTG) trace contains fetal heart rate (FHR, upper) and uterine contraction (UC, lower) data. The trace is rendered at a paper speed of 1 cm/min. \protect\B{} Five asynchronous Feature Agents \protect\asmall{}-\protect\esmall{} evaluate \textit{Baseline}, \textit{Variability}, \textit{Accelerations}, \textit{Decelerations}, and \textit{Sinusoidal Pattern} of the CTG respectively. Each agent outputs a classification and interpretation of the corresponding feature. \protect\C{} Once all Feature Agents have completed analyses, the results are aggregated and fed into the Aggregator Agent. This agent provides an overall classification of the CTG and generates a comprehensive explanation based on the assessments of all features.}
    \label{fig:system_architecture}
\end{figure*}

Cardiotocography (CTG) plays a central role in prenatal care by continuously monitoring fetal heart rate (FHR) and uterine contraction (UC), enabling early detection of potential fetal distress. With the increasing adoption of remote fetal monitoring technologies, many intrapartum pregnancies use home-use CTG devices connected to applications to collect CTG data \cite{van2020home}. In most remote monitoring scenarios, the data are either presented as raw numerical readings or visual CTG traces (See Figure~\ref{fig:system_architecture}\A{}), without medical context or actionable interpretation. Clinicians, due to workload or asynchronous access, often cannot provide immediate responses \cite{li2022latent}. As a result, users are left to interpret complex physiological signals on their own, leading to confusion, frustration, or even anxiety, especially when faced with ambiguous patterns or unexplained alerts. This disconnect highlights a broader challenge in the design of human-centered medical technologies: how can we provide not just accurate detection, but understandable and transparent explanations \cite{shneiderman2020bridging, shneiderman2022human, shneiderman2020human} that support both clinical decision-making and user comprehension? While existing deep learning models have achieved promising results in CTG classification, they often operate as black boxes, providing little insight into why a particular diagnosis is made. Such opacity limits user trust, interpretability, and real-world adoption.

To address this gap, we present \textbf{\textit{CTG-Insight}}, a multi-agent, LLM-based system that performs interpretable analysis of CTG data. Rather than treating the task as an end-to-end black-box classification problem, CTG-Insight decomposes the interpretation into five medically defined features: baseline, variability, accelerations, decelerations, and sinusoidal pattern, each analyzed by a dedicated agent. These agents classify their respective features as normal, suspicious, or pathological, and provide natural language explanations based on clinical guidelines. A final aggregation agent synthesizes these results to produce an overall assessment of fetal status, accompanied by a structured explanation. We evaluate the system on a publicly available dataset, comparing CTG-Insight with traditional deep learning models and an LLM-based baseline. Results show that our system achieves state-of-the-art classification accuracy while generating interpretable, structured explanations that follow clinical reasoning patterns. The contributions of this paper are threefold: (1) We perform a user need analysis of existing fetal monitoring apps and identify key gaps in interpretability and emotional support that motivate our system design. (2) We propose CTG-Insight, a modular multi-agent LLM framework that mirrors clinical reasoning and provides feature-level explainability for CTG interpretation. (3) We align all agent prompts with established clinical guidelines to ensure medically grounded interpretations and improve the trustworthiness of model outputs. This work lays the foundation for future development of integrated CTG devices and mobile applications that incorporate our system, with the long-term goal of enhancing emotional reassurance, clinical communication, and user-centered feedback in intrapartum care. We situate our contributions within broader discussions in ubiquitous computing, human-computer interaction, wellbeing, and health monitoring.


\section{Related Work}

\subsection{XAI and User-Centered Design in Health Monitoring}

Recent years have seen increasing interest in applying machine learning to CTG classification~\cite{sun2024neurofetalnet, liang2023cnn, he2016deep}, yet these models often lack transparency, raising concerns about their applicability in clinical or patient-facing contexts. From a human-computer interaction (HCI) perspective, explainability and user-centeredness are critical in the design of health technologies~\cite{chen2022explainable, liang2021human, schemmer2023appropriate, naiseh2023different}, especially in sensitive domains such as prenatal care. Work by Binns et al.~\cite{binns2018s} and Eiband et al.~\cite{eiband2018bringing} emphasizes that explanations must be actionable and sensitive to user emotional states—key concerns for expectant parents interpreting fetal monitoring data. In remote health monitoring, Johnson et al.~\cite{johnson2005user} and Tutty et al.~\cite{tutty2019complex} have shown that raw data displays often overwhelm users, who need interpretive and contextual feedback rather than mere signal exposure. Furthermore, studies on trust in algorithmic systems~\cite{bussone2015role, doshi2017towards, yin2019understanding} highlight the importance of making system reasoning visible and relatable. In explainable AI (XAI), approaches such as LIME and SHAP have been proposed to provide post-hoc interpretability for black-box models~\cite{ribeiro2016should, lundberg2017unified}. However, such methods are typically designed for general machine learning settings and may not align well with domain-specific, clinically guided reasoning required in healthcare. By contrast, rule-based and inherently interpretable models—such as decision trees or case-based reasoning—have been used in clinical decision support systems (CDSS) to improve transparency~\cite{antoniadi2021current, shortliffe1974rule}. Still, these often sacrifice performance or flexibility. Meanwhile, large language models (LLMs) have emerged as powerful tools for generating natural language explanations in medicine~\cite{hu2024exploring, tu2024towards, he2025survey}. Yet, few systems have structured these capabilities around clinically validated frameworks for physiological signal interpretation. Our work bridges these threads by designing an interpretable, LLM-driven CTG analysis system tailored to remote fetal monitoring users' informational and emotional needs, while grounding its reasoning in clinical rules and domain knowledge.


\subsection{Formative Analysis of Existing Apps}


To inform our system design, we conducted a formative analysis of some widely available fetal heart monitoring apps on the Chinese market:
\textit{Xiya} (\begin{CJK}{UTF8}{gbsn}喜芽\end{CJK}),
\textit{Mengdong} (\begin{CJK}{UTF8}{gbsn}萌动\end{CJK}),
\textit{Weitaixin} (\begin{CJK}{UTF8}{gbsn}微胎心\end{CJK}),
\textit{Fetalheart}, and
\textit{Taixinbao} (\begin{CJK}{UTF8}{gbsn}胎心宝\end{CJK}). These apps were selected for their popularity and representation across domestic software platforms. Our analysis focused on features offered, interaction design, interpretability of information, and support for users’ emotional needs. While all apps supported basic FHR (only a few support UC) monitoring, Doppler audio playback, and historical data charts, we observed several common limitations that shaped our design goals: (1) \textbf{Lack of interpretability:} FHR was typically presented as raw numbers or simple charts without contextual explanation. Users often complained about whether values were normal or concerning, and apps provided no guidance. (2) \textbf{Missing emotional support:} None of the apps included empathetic language, affective feedback, or any support for the emotional well-being of pregnant users, despite pregnancy being a time of elevated anxiety and need for reassurance.


\section{Methodology}


\subsection{Dataset Selection}
We considered several public CTG datasets, including the UCI CTG Dataset~\cite{cardiotocography_193}, the CTU-UHB Intrapartum CTG Database~\cite{Chudcek2014OpenAI}, and the NeuroFetalNet Dataset~\cite{sun2024neurofetalnet}. To support CTG trace reconstruction and feature analysis, we required continuous FHR and UC signals from antepartum pregnancies. The first two datasets did not meet these criteria, so we selected NeuroFetalNet. It provides 20-minute CTG traces sampled at 4~Hz (4,800 points) with clinician-annotated binary labels (normal/abnormal, with suspicious cases categorized as abnormal).

\subsection{System Design and Implementation}

\subsubsection{CTG-Insight Workflow Design}
The design of the CTG-Insight system is inspired by common deep learning architectures, which typically consist of a feature extraction phase (input and hidden layers) followed by a classification phase (output layer) \cite{krizhevsky2017imagenet}. However, the feature extraction process in deep learning models is often non-interpretable, making it challenging to explain decisions \cite{samek2017explainable}. To address this limitation, the CTG-Insight multi-agent system is structured into two distinct components: (1) \textbf{Feature Agents} (see Figure~\ref{fig:system_architecture}\B{}): These agents emulate the feature extraction process in deep learning. Five agents operate in parallel, each analyzing the same CTG trace but focusing on a specific feature. A customized prompt guides each agent and independently produces a classification (normal, suspicious, or pathological) along with an explanation for its decision. (2) \textbf{Aggregator Agent} (see Figure~\ref{fig:system_architecture}\C{}): This agent simulates the role of the output layer in a deep learning model. It aggregates the outputs from all Feature Agents and synthesizes the information based on a predefined prompt to generate an overall classification of fetal health, along with a comprehensive explanation.

\subsubsection{Prompt Design}
Based on the FIGO Guidelines \cite{ayres2015figo} and the German S1-Guideline \cite{schneider2014s1}, we identified five key features for CTG analysis: Baseline, Variability, Accelerations, Decelerations, and Sinusoidal Pattern. For each feature, we established criteria for classifying the signal as normal, suspicious, or pathological. These classification criteria were adapted from the official guidelines, with slight modifications to account for the limited 20-minute duration of each data sample in the selected dataset. As shown in Appendix~\ref{baseline}-\ref{sinusoidal}, each Feature Agent is guided by a prompt that includes (1) \textbf{Definition:} the definition of the feature, (2) \textbf{Rule:} the classification criteria (normal/suspicious/pathological), (3) \textbf{Role:} the agent’s role in performing classification and providing explanations, and (4) \textbf{Example Output:} three example output to ensure consistent formatting. In particular, the prompt for the Decelerations Agent (see Appendix~\ref{decelerations}) includes an additional section, (5) \textbf{Type:} the types of various decelerations. This is necessary because decelerations encompass a wide range of patterns, and the guidelines specify classification rules based on these types; therefore, we explicitly incorporated type definitions into the prompt to support accurate recognition and interpretation. For the Aggregator Agent, the prompt followed the guidelines for synthesizing individual feature assessments into an overall judgment. This agent is instructed to summarize the fetal condition as normal, suspicious, or pathological, and to explain the reasoning behind its final decision (see Appendix~\ref{overall}). We developed the system using Python, visualized CTG traces from time series data using the Matplotlib library, and built all agents using OpenAI’s GPT-4.1 model\footnote{\url{https://platform.openai.com/docs/models/gpt-4.1}}.

\section{Experiments}

\subsection{Experiments Settings}
To evaluate the effectiveness of CTG-Insight, we first aimed to assess its accuracy. We divided the comparison into two categories. The first category involved LLM-based methods, which leveraged both textual and visual capabilities of the LLM. The input consisted of structured prompts and CTG traces generated by visualizing the time-series FHR and UC data from the NeuroFetalNet Dataset. Within this category, we compared our CTG-Insight system against a baseline method referred to as Direct Prompt, in which a single agent receives a concatenated set of all prompts from Appendix~\ref{overall}-\ref{sinusoidal} and directly outputs the final classification without decomposing the task into feature-level analysis. The second category included deep learning-based methods, specifically Sun et al.'s NeuroFetalNet \cite{sun2024neurofetalnet}, the CNN+BiGRU model used by Liang et al. \cite{liang2023cnn}, and the ResNet-based approach \cite{sun2024neurofetalnet, he2016deep}. Since the NeuroFetalNet Dataset consists of time-series data and is already partitioned into training and testing sets in a 9:1 ratio, we trained and tested the deep learning models directly on the raw time-series input. For the LLM-based methods, we randomly sampled 50 instances (25 labeled as abnormal and 25 as normal) from the test set and visualized the time-series data into CTG traces for prediction. We used Accuracy and F1-score as evaluation metrics, and all results reported are the average values over five repeated trials.

\subsection{Results and Discussion}
Table~\ref{tab:performance_comparison} presents the performance comparison among different methods evaluated on the NeuroFetalNet Dataset. Our proposed CTG-Insight system outperformed all baselines, achieving an Accuracy of 96.40\% and an F1-score of 97.81\%, substantially surpassing both LLM-based and deep learning-based alternatives. Among the LLM-based methods, Direct Prompt, where a single agent receives all prompts and produces a final classification, achieved an Accuracy of 79.80\% and an F1-score of 80.10\%. One plausible explanation for the relatively poor performance of the Direct Prompt baseline is the tendency of LLMs to overlook or forget specific instructions when presented with excessively long and complex prompts \cite{kusano2024longer, agrawal2024can}. By assigning dedicated agents to analyze individual CTG features, CTG-Insight produced more accurate and consistent classifications, benefiting from structured reasoning and modular decision-making. Compared to deep learning models, CTG-Insight also achieved superior performance. While NeuroFetalNet~\cite{sun2024neurofetalnet} performed strongly with an Accuracy of 94.23\%, CTG-Insight still achieved higher scores on both metrics. These results highlight several key advantages of CTG-Insight. First, it leverages domain-specific knowledge explicitly encoded into structured prompts, improving transparency and performance. Second, the multi-agent framework promotes modular interpretability, enabling clear attribution of predictions to specific CTG features. Finally, when guided by domain-aligned prompts and visual inputs, the strong generalization of LLMs allows for high accuracy even in small-sample scenarios, where traditional models may underperform.

\begin{table}[htbp]
\centering
\caption{Performance Comparison of Different Methods on the NeuroFetalNet Dataset}
\label{tab:performance_comparison}
\begin{tabular}{lcc}
\toprule
\textbf{Method} & \textbf{Accuracy (\%)} & \textbf{F1-Score (\%)} \\
\midrule
\textbf{LLM-based Methods} & & \\
\quad CTG-Insight (Ours) & \textbf{96.40} & \textbf{97.81} \\
\quad Direct Prompt & 79.80 & 80.10  \\
\midrule
\textbf{Deep Learning Methods} & & \\
\quad NeuroFetalNet \cite{sun2024neurofetalnet} & 94.23  & 94.20  \\
\quad CNN+BiGRU \cite{liang2023cnn} & 84.04 & 84.16  \\
\quad ResNet \cite{sun2024neurofetalnet,he2016deep} & 82.88  & 82.79 \\
\bottomrule
\end{tabular}
\end{table}

\section{Limitations and Future Work}

While this work demonstrates the feasibility and interpretability of CTG-Insight, several limitations remain that suggest directions for future research. First, we have not yet conducted user studies to assess how stakeholders perceive and engage with the system's explanations, including expectant mothers, clinicians, and midwives. Future work will involve stakeholder-centered evaluations through semi-structured interviews and quantitative assessments (e.g., Likert-scale surveys) to measure interpretability, trust, and emotional reassurance. Second, CTG-Insight currently operates in an offline setting using pre-recorded CTG traces. However, real-world fetal monitoring is continuous and time-sensitive. We plan to extend CTG-Insight for real-time streaming data by re-architecting agents for continuous input and state tracking. This will also require integrating mobile or wearable platforms for dynamic use in home or clinical environments. Third, while our current design aligns with clinical guidelines, it does not yet account for contextual factors such as gestational age, maternal history, or labor stage, which are critical for nuanced decision-making. Future iterations will incorporate such metadata to support personalized interpretations and improve clinical relevance. Finally, we acknowledge that the 50-sample subset used in LLM evaluation may limit generalizability and statistical significance. A full-scale evaluation is planned as future work. Overall, this work lays a foundation for interpretable, guideline-driven CTG analysis. Future development will focus on real-time deployment, contextual modeling, and validation with end-users to enhance system robustness and usability in real-world scenarios.




\clearpage

\bibliographystyle{ACM-Reference-Format}
\balance
\bibliography{reference}

\appendix
\onecolumn
\section{System Prompt}

\subsection{Overall CTG Analysis Prompt}\label{overall}
\begin{verbatim}
The CTG classification falls into one of the following three categories:
- Normal: All evaluated features are within normal ranges. No suspicious or pathological findings are present.
- Suspicious: One feature is classified as suspicious while all others remain normal.
- Pathological:(1) At least one feature is pathological; or (2) At least two features are suspicious.

(Direct Prompt) Based on the definition and rules, please classify the CTG image into one of the following 
categories: Normal, Suspicious, Pathological, and provide a thorough explanation for your classification.

(CTG-Insight) Based on all the features' analysis and classifications, please provide a final classification 
and a thorough explanation of the CTG tracing.

\end{verbatim}

\subsection{Baseline Analysis Prompt}\label{baseline}
\begin{verbatim}
# Baseline_Definition
- Baseline is the mean FHR maintained over at least 10 minutes **in the absence of accelerations or decelerations**.
- The baseline value may vary between subsequent 10-minute sections.

# Baseline_Rule
- Normal: 110–160 bpm
- Suspicious: 100–109 bpm, or 161–180 bpm
- Pathological: <100 bpm or >180 bpm
- **Note**: If persistent fluctuations occur, such as sustained prolonged decelerations or accelerations, 
and the baseline cannot be determined, the pattern should be considered Suspicious.

# Role
You are an expert in CTG analysis. You will be given a CTG tracing and you need to classify the Baseline of 
the tracing into one of the following categories: Normal, Suspicious, Pathological. 
And you need to provide a brief explanation for your classification.
(All roles in the prompt are similar, so we omit the following for space-saving purposes.)

# Example Output
(We omit all example entries in the prompt for space-saving purposes.)
\end{verbatim}

\subsection{Variability Analysis Prompt}
\begin{verbatim}
# Variability_Definition
- Variability is the degree of change in FHR over time.
- It is evaluated as the average bandwidth amplitude of the signal in 1-minute segments.
- Fluctuations in the fetal baseline rate that occur **3–5 times per minute** are considered normal.

# Variability_Rule
- Normal: 5–25 bpm, 3-5 times per minute, appears in most of the time (both frequency and amplitude criteria 
  must be simultaneously fulfilled.)
- Suspicious: Mildly abnormal but does not meet pathological thresholds (e.g., <5 bpm lasting >=15 minutes, 
  or >25 bpm lasting >=10 minutes)
- Pathological: <5 bpm lasting >=15 minutes, or >25 bpm lasting >=10 minutes
\end{verbatim}

\subsection{Accelerations Analysis Prompt}
\begin{verbatim}
# Accelerations_Definition
- Accelerations are defined as increases in the FHR **above the baseline**, of more than 15 bpm in amplitude, 
and lasting more than 15 seconds.
- **Note**: **Baseline** is the mean FHR maintained over at least 10 minutes **in the absence of accelerations 
or decelerations**.

# Accelerations_Rule
- Normal: two accelerations in 20minutes
- Suspicious: periodical occurrence with every contraction
- Pathological: no accelerations
\end{verbatim}

\subsection{Decelerations Analysis Prompt}\label{decelerations}
\begin{verbatim}
# Decelerations_Definition
- Decelerations are defined as decreases in the FHR **below the baseline**, of more than 15 bpm in amplitude, 
  and lasting more than 15 seconds..
- **Note**: **Baseline** is the mean FHR maintained over at least 10 minutes **in the absence of accelerations 
  or decelerations**.

# Decelerations_Type
- Early decelerations: Shallow drops in heart rate (usually >15 bpm and >15 seconds) mirror uterine 
  contractions. They are caused by fetal head compression and are considered benign.
- Variable decelerations: Abrupt, V-shaped drops (>15 bpm drop within <30 seconds, lasting >15 seconds), 
  often linked to cord compression, typically harmless unless prolonged (>3 min) or with reduced variability.
- Late decelerations: Gradual, U-shaped declines and/or with reduced variability in heart rate (onset to 
  nadir >30 seconds), beginning >20 seconds after contraction starts and returning after it ends, 
  indicating uteroplacental insufficiency and potential fetal hypoxia—especially concerning when variability 
  is reduced or amplitude is only 10–15 bpm.
- Prolonged decelerations: Sustained drops in heart rate (>15 bpm lasting >3 minutes), and when exceeding 5 minutes
  with FHR <80 bpm and reduced variability, they signal acute fetal hypoxia/acidosis requiring urgent intervention.
- Atypical Variable Decelerations: decelerations with one of the following additional characteristics:
    - loss of primary or secondary FHR rise
    - slow return to baseline after the contraction has ended
    - longer increased baseline after contraction
    - biphasic deceleration
    - loss of oscillation during deceleration
    - resumption of baseline rate at a lower level

# Decelerations_Rule
- Normal: No decelerations
- Suspicious: Early decelerations, Variable decelerations, Prolonged decelerations(<3min)
- Pathological: Late decelerations, Prolonged decelerations(persist for more than two contractions or >3minutes), 
  Atypical Variable Decelerations
\end{verbatim}

\subsection{Sinusoidal Pattern Analysis Prompt}\label{sinusoidal}
\begin{verbatim}
# Sinusoidal_Definition
- A regular, smooth, undulating signal, resembling a sine wave, with an amplitude of 5-15bpm, and a frequency 
  of 3-5 cycles per minute.
- This pattern lasts more than 10 minutes, and coincides with absent accelerations.

# Pseudosinusoidal_Definition
- A pattern resembling the sinusoidal pattern, but with a more jagged “saw-tooth” appearance, rather than the 
  smooth sine-wave form.
- It is characterized by normal patterns before and after.

# Sinusoidal_Rule
- Normal: No sinusoidal pattern
- Suspicious: Pseudosinusoidal pattern (lasting <10 minutes, with atypical morphology)
- Pathological: True sinusoidal pattern (smooth sine-wave, amplitude 5–15 bpm, frequency 3–5 cycles/min, 
  lasting >=10 minutes, accompanied by absent accelerations)
\end{verbatim}

\end{document}